# Towards Interpretable Renal Health Decline Forecasting via Multi-LMM Collaborative Reasoning Framework

Peng-Yi Wu
*Department of Information Management*
*National Sun Yat-Sen University*
Kaohsiung, Taiwan
pengyiwu0963@gmail.com

Pei-Cing Huang
*Department of Information Management*
*National Sun Yat-Sen University*
Kaohsiung, Taiwan
pcpeicing@gmail.com

Ting-Yu Chen
*Department of Information Management*
*National Sun Yat-Sen University*
Kaohsiung, Taiwan
brontotingyu@gmail.com

Chantung Ku
*Department of Information Management*
*National Sun Yat-Sen University*
Kaohsiung, Taiwan
kuchantung@gmail.com

Ming-Yen Lin
*Kaohsiung Medical University Hospital*
*Kaohsiung Medical University*
Kaohsiung, Taiwan
mingyenlin3@gmail.com

Yihuang Kang
*Department of Information Management*
*National Sun Yat-Sen University*
Kaohsiung, Taiwan
ykang@mis.nsysu.edu.tw

*Abstract*—Accurate and interpretable prediction of estimated glomerular filtration rate (eGFR) is essential for managing chronic kidney disease (CKD) and supporting clinical decisions. Recent advances in Large Multimodal Models (LMMs) have shown strong potential in clinical prediction tasks due to their ability to process visual and textual information. However, challenges related to deployment cost, data privacy, and model reliability hinder their adoption. In this study, we propose a collaborative framework that enhances the performance of open-source LMMs for eGFR forecasting while generating clinically meaningful explanations. The framework incorporates visual knowledge transfer, abductive reasoning, and a short-term memory mechanism to enhance prediction accuracy and interpretability. Experimental results show that the proposed framework achieves predictive performance and interpretability comparable to proprietary models. It also provides plausible clinical reasoning processes behind each prediction. Our method sheds new light on building AI systems for healthcare that combine predictive accuracy with clinically grounded interpretability.

## I. INTRODUCTION

Chronic kidney disease (CKD) is a growing global health concern, with prevalence estimates ranging from 3% to 18% of the population [1]. The estimated glomerular filtration rate (eGFR) is an essential indicator for diagnosing CKD and detecting early kidney damage. Accurate eGFR prediction can support early intervention and improve clinical outcomes [2]. While traditional formulas, such as CKD-EPI [3], are widely used to estimate kidney function, forecasting future kidney function decline is still a challenge. Recent machine learning (ML) models have shown promise in improving eGFR prediction, but many still fall short in capturing patient-specific trends and often lack interpretability. For AI models to be adopted in clinical settings, transparency and explainability are essential [4]. Clinicians must be able to understand how predictions are made to build trust and identify potential errors. Therefore, predictive models should not only be accurate but also provide explanations that reflect clinical reasoning and are easy to interpret.

Large Language Models (LLMs) and Large Multimodal Models (LMMs) demonstrated remarkable capabilities in processing both language and multimodal data [5]. These models, pretrained on large-scale datasets, have shown potential in medical applications by integrating domain knowledge and generating interpretable outputs [6]. Their ability to process text and visual data opens new opportunities for improving eGFR prediction and explanation quality. Recent work has demonstrated that API-based, commercial LMMs can outperform traditional ML models in eGFR prediction when provided with proper prompts [7]. However, concerns over data privacy and high implementation costs limit their clinical use. In contrast, open-source and open-weight LMMs are more suitable for local deployment but often struggle with complex clinical reasoning and visual interpretation, which may lead to unreliable or hallucinated outputs [8]. These limitations highlight the need to develop a more accurate and interpretable prediction framework to improve these shortcomings.

To address these issues, this study proposes a collaborative multi-LMM framework designed to improve the predictive performance of open-source LMMs in eGFR forecasting while providing clinically meaningful explanations. Drawing inspiration from the concept of Social Learning [9], the framework facilitates knowledge transfer from stronger to smaller models, improving their ability to interpret visuals and generate accurate predictions. To support transparent and clinically aligned reasoning, we integrate Chain-of-Thought (CoT) [10] reasoning to support step-by-step inference and introduce a short-term memory mechanism to enhance consistency and refinement across sequential predictions. We further adopt abductive reasoning [11] to generate data-driven and hypothesis-based explanations that offer objective perspectives and support comprehensive clinical interpretations. The key contributions of this study are summarized as follows:

- We propose an LMM prediction framework combining explanatory and causal reasoning for renal function forecasting.
- We explore and compare multiple techniques to enhance the performance of open-source models in eGFR prediction tasks.

- We incorporate structured abductive reasoning to generate clinical explanations that go beyond surface-level data patterns and offer reasoning processes that are more compatible with clinical thinking.

The rest of this paper is organized as follows: Section 2 reviews existing eGFR prediction methods and recent advances in explainable and reasoning-enhanced AI approaches. Section 3 outlines our proposed framework, followed by Section 4, which presents the results of our experiments, comparing our approach to conventional ML models and different types of LMMs. Section 5 discusses the limitations of the current study and outlines directions for future research. Finally, Section 6 concludes the paper by summarizing the main contributions and practical implications of the proposed framework.

## II. BACKGROUND

CKD is the seventh leading cause of death worldwide, with mortality increasing by 50% from 2000 to 2019 and estimated to account for 5 to 11 million deaths annually [12]. The eGFR is the primary clinical indicator of kidney function, playing a critical role in diagnosing CKD. Derived from serum creatinine and demographic variables such as age and sex, eGFR calculated by formula provides a standard measure of renal function. According to the National Kidney Foundation, regular eGFR monitoring is essential for individuals at risk, including those with diabetes, hypertension, or a family history of kidney disease [13].

Traditional formulas estimate eGFR using static variables. However, they are limited in capturing dynamic changes and are influenced by individual differences in muscle mass, diet, and comorbidities. More recently, ML models have been introduced to enhance eGFR prediction accuracy by incorporating multivariate and temporal data. Prior studies have demonstrated the utility of ML in predicting eGFR through ultrasound-based kidney imaging, and the results outperformed traditional methods and were more accurate than most nephrologists' assessments [14]. Another utilized a random forest (RF) model to predict the progression of CKD using demographics and laboratory data [15]. While these models offer potential for improved accuracy, several limitations remain. Many suffered from limited generalizability due to training on single-center or region-specific datasets. Furthermore, their performance is often dependent on large data volumes, and interpretability remains a key barrier, particularly for deep learning models, which are often perceived as "black boxes", making it challenging for clinicians to understand prediction rationales.

Model transparency is particularly important in healthcare, where predictive outputs influence real-world medical decisions [4]. This has led to growing interest in explainable AI (XAI) frameworks [16]. Techniques such as SHAP [17] and LIME [18] have been widely used to provide explanations by attributing model outputs to input features. A recent study further emphasizes the need for patient-specific interpretations to support precise, individualized predictions, particularly in practical applications where understanding the rationale behind a single patient's outcome is more critical than identifying general population trends [19]. While these techniques improve interpretability, many existing interpretable methods still rely on surface-level correlations. Furthermore, they are limited in their ability to produce narrative, context-aware explanations that align with clinical reasoning.

The advancements in LLMs and LMMs have opened new opportunities for both predictive accuracy and interpretability in clinical tasks. For instance, Hegselmann et al. used LLMs to classify tabular healthcare data by serializing rows into natural-language prompts [20]. However, performance may suffer when input features exceed token limits. Our previous work investigated modality transformation, where tabular clinical records are converted into visual formats that LMMs can interpret more effectively [7]. This approach showed that commercial models, such as Google Gemini [21] and GPT-4o [22], achieve superior performance in eGFR prediction when given visualized input and proper prompts.

Despite their strong performance, commercial LMMs raise concerns regarding data privacy, cost, and limited local control. In contrast, open-source LMMs offer the advantage of local deployment and enhanced data governance but often struggle with complex reasoning and visual interpretation [23], increasing the risk of hallucinated outputs [8]. Inspired by how humans learn from one another, Scao et al. introduced the idea of social learning among LLMs, where models exchange knowledge using natural language [9]. This approach suggests that LLM-based agents could benefit from interacting and learning collaboratively to further enhance their performance. In parallel, memory mechanisms have been proposed to help models retain contextual information across time, improving consistency and reasoning in sequential clinical tasks [24].

Building on these developments, recent research has also explored ways to equip LLMs with human-like reasoning abilities. Among them, CoT prompting is recognized as a representative method that encourages models to articulate intermediate reasoning steps in natural languages [10]. This approach enhances the model's capacity for performing complex reasoning tasks. Formal reasoning paradigms, such as inductive, deductive, and abductive reasoning, have also gained attention. Abductive reasoning is considered a valuable cognitive tool in clinical decision-making, particularly under uncertainty, as it supports individualized judgments by bridging population-level evidence and patient-specific contexts [25]. However, the use of abductive reasoning to generate explanations remains relatively underexplored in the field of AI in medicine. Simulating plausible clinical reasoning may offer value by enhancing the interpretability of model outputs and serving as a pedagogical tool for training medical professionals. Therefore, our work explores techniques such as knowledge transfer and memory mechanisms to enhance the prediction accuracy of open-source LMMs for eGFR forecasting while generating interpretable and clinically meaningful explanations through abductive reasoning. We aim to enable more trustworthy, transparent, and locally deployable AI systems for practical healthcare use.

## III. INTERPRETABLE TEACHER-STUDENT LMM FRAMEWORK FOR eGFR FORECASTING

We present our proposed framework, which aims to improve open-source LMM eGFR prediction accuracy while providing interpretable reasoning behind the forecasts. We integrate two key components: Knowledge Transfer, which allows the model to learn visual understanding from stronger vision-language models, and Short-term Memory, which enhances contextual continuity in reasoning and explanation generation. The framework, as illustrated in Figure 1, consists of two stages: image interpretation and eGFR prediction with explanation.

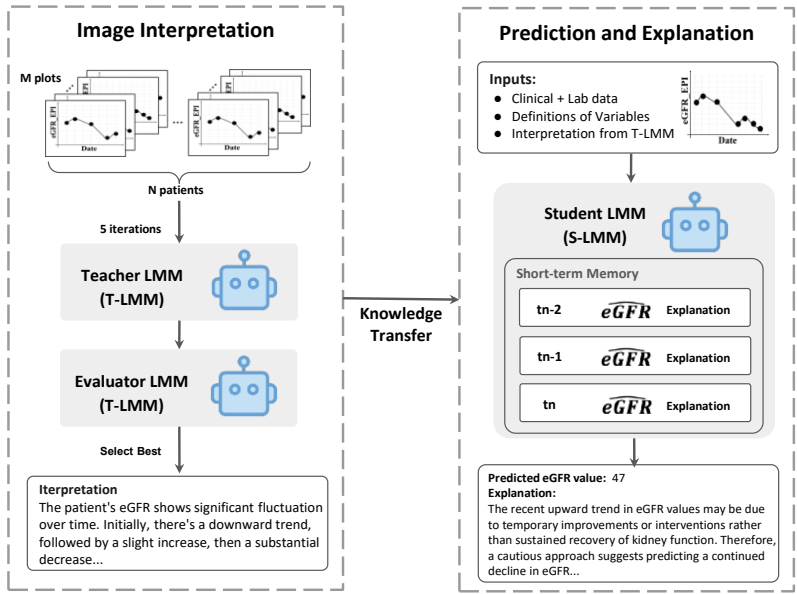

Fig. 1. Overview of the proposed two-stage eGFR prediction.

In the initial phase, each patient's historical eGFR measurements are first transformed into a series of de-identified trend line charts to preserve privacy. These charts are input sequentially into a proprietary vision-language model, also called Teacher LMM (T-LMM), to extract clinically meaningful trends and summarize kidney function status. For each patient, we generate a sequence of $M$ line charts, where $M$ is determined by the median number of available eGFR measurements. The patient's m-th plot includes one additional data point compared to the (m–1)-th plot, specifically the next eGFR value and its corresponding data to support progressive interpretation. The T-LMM produces structured interpretations for each chart, which are then evaluated by the same model, acting as an Evaluator LMM (E-LMM), using a predefined rubric to assess clinical accuracy and coherence. The highest-scoring interpretation is selected as the final output. Through this process, we obtain image-derived textual summaries that serve as external knowledge for the next stage of prediction.

In the second stage, we leverage the Student LMM (S-LMM), an open-source language model that is deployed locally, to generate future eGFR predictions and clinical explanations. The model is given multiple inputs, including the patient's eGFR trajectory in chart format, structured clinical and laboratory variables, and T-LMM-generated interpretations, which together provide comprehensive contexts for the prediction task. We implement a CoT prompting strategy to enhance its reasoning capabilities and reduce the risk of hallucination. Specifically, the S-LMM first predicts the next eGFR value and then generates an explanation based on the predicted output. This design ensures that explanations are logically grounded in the model's actual prediction, reducing inconsistencies or fragmented responses. By anchoring the reasoning process to the prediction step, the model can explain how specific trends or variables influence the predicted outcome, mimicking clinical reasoning.

Since retaining contextual information over time is crucial for coherent reasoning in sequential clinical tasks, we also incorporate a short-term memory mechanism to improve consistency across prediction steps. After each prediction, the prompt, predicted value, and explanation are stored. When the model proceeds to the next step, it retrieves this memory along with the ground truth of the prior prediction, allowing it to self-correct and refine its reasoning. This mechanism reinforces temporal continuity and encourages the model to recognize evolving patterns across time points, thereby improving predictive accuracy.

To enhance clinical interpretability, the model generates structured explanations through two complementary forms of abductive reasoning [25]: selective abduction, which grounds predictions in observed clinical data [e.g., eGFR trends, blood urea nitrogen (BUN), urine albumin creatinine ratio (UACR)], and creative abduction, which hypothesizes plausible but unobserved factors that could contribute to the predicted outcome. By combining data-driven and hypothesis-based reasoning, the model can generate more comprehensive interpretations and may help mitigate diagnostic bias. It also shows potential as an educational tool for training students and junior clinicians in structured clinical reasoning.

## IV. EXPERIMENTS AND DISCUSSION

To demonstrate our proposed method, we utilize data from the Kaohsiung Medical University Research database. The dataset contains comprehensive variables, including demographic information, clinical biomarkers, and lifestyle factors. Participants were outpatients who received regular follow-up care from nephrologists at two affiliated hospitals of Kaohsiung Medical University between 2004 and 2021. This study was approved by the Institutional Review Board of Kaohsiung Medical University Hospital (KMUHIRB-EXEMPT(I)-20210123). For simplicity, we focus on a subset of 570 observations, randomly selected from 50 patients with varying baseline eGFR levels. We divided the patients into 70% for training and 30% for validation, while maintaining a similar distribution of CKD stages across both sets to ensure fair evaluation across disease severity levels.

TABLE I. COMPARISON OF MODEL PERFORMANCE WITH DIFFERENT METHODS.

| Model | Method | Train | | Validation | |
|---|---|---|---|---|---|
| | | MAE | MAPE(%) | MAE | MAPE(%) |
| RF | - | 1.02 | 4.35 | 2.64 | 11.06 |
| 1D-CNN | - | 2.76 | 12.06 | 2.60 | 11.74 |
| Llama 3.2 vision 11B | zero-shot | 4.76 | 19.18 | 4.33 | 18.68 |
| | knowledge transfer | 3.81 | 14.86 | 4.06 | 31.00 |
| | knowledge transfer + short-term memory | 3.62 | 13.53 | 3.87 | 24.54 |
| Gemma 3 12B | zero-shot | 4.80 | 18.60 | 4.16 | 17.10 |
| | knowledge transfer | 3.76 | 14.98 | 3.78 | 26.90 |
| | knowledge transfer + short-term memory | 3.73 | 15.24 | 3.84 | 28.34 |
| Qwen 2.5 vision 32B | zero-shot | 2.99 | 12.46 | 2.76 | 12.29 |
| | zero-shot + short-term memory | 4.10 | 14.96 | 3.46 | 15.00 |
| GPT-4o | zero shot | 2.57 | 11.66 | 2.50 | 11.72 |
| | zero-shot + short-term memory | 2.55 | 11.53 | 2.55 | 11.51 |
| Gemini 1.5 Pro | zero-shot | 3.20 | 13.14 | 2.82 | 12.28 |
| | zero-shot + short-term memory | 3.13 | 12.91 | 2.79 | 12.30 |
| Gemini 2.0 Pro | zero-shot | 3.21 | 14.23 | 3.12 | 14.37 |
| | zero-shot + short-term memory | 3.75 | 15.40 | 3.24 | 15.37 |
| Gemini 2.0 Flash | zero-shot | 2.64 | 11.90 | 2.65 | 11.78 |
| | zero-shot + short-term memory | 2.86 | 12.92 | 2.45 | 11.17 |
| Claude 3.7 Sonnet (self-moderated) | zero-shot | 2.98 | 12.86 | 2.94 | 12.46 |
| | zero-shot + short-term memory | 2.98 | 13.07 | 2.69 | 13.27 |

We began our experiment by converting each patient's historical eGFR measurements into line charts, where the X-axis represents the dates of each eGFR measurement, and the Y-axis displays the eGFR values in units of mL/min/1.73m². We use Gemini 1.5 Pro [26] as the T-LMM to interpret the charts and generate structured visual summaries, which are later provided to the S-LMM for prediction tasks. T-LMM extracts clinically relevant features from the plots, including identifying inflection points, assessing recent trends, estimating short-term changes, and classifying kidney status based on CKD staging criteria. Following the previous stage, we input patients' line charts, clinical and laboratory data, and interpretations from T-LMM into S-LMMs, such as Llama 3.2 [27], [28] and Gemma 9b [29], to predict eGFR. Before generating the final prediction, the S-LMM is first prompted to predict the two most recent eGFR values. This step helps the model gradually construct an understanding of the patient's trajectory. The interactions from these two predictions, including prompts, outputs, and explanations, are stored in S-LMM's short-term memory to support contextual continuity. The model followed a step-wise process, first generating predictions and then providing clinical explanations corresponding to its predictions. This sequential chaining approach ensures the explanation is grounded in the model's prediction, reducing the risk of fragmented or incomplete responses.

We evaluate two types of LMMs on our framework: (1) open-source LMMs, including Llama 3.2 vision 11b and Gemma 3 12b; (2) proprietary models, including GPT-4o [22], Gemini 1.5 pro [26]. We also compare our framework to traditional models, including a Random Forest (RF) and a one-dimensional convolutional neural network (1D-CNN), which are only provided with clinical and variable data for predictions. Prediction accuracy is assessed using Mean Squared Error (MSE) and Mean Absolute Percentage Error (MAPE).

Our framework is designed to be modular and adaptable, allowing different components to be flexibly combined based on model capabilities. For instance, models with limited visual-language integrated capabilities can benefit from knowledge transfer mechanisms. In contrast, models that already possess sufficient visual understanding may not require such assistance and can instead benefit from a short-term memory mechanism. To assess the effectiveness of each design choice, we compare configurations with and without specific components, such as knowledge transfer and memory mechanisms, across models with varying capacities. As shown in Table 1, the best overall performance in terms of MAE and MAPE was achieved by the RF model. However, the model does not provide interpretability or explanatory outputs, which limits its clinical applicability. Furthermore, RF and 1D-CNN require training on the entire dataset, whereas LLMs and our framework utilize in-context learning. This allows the model to make predictions based only on inference-time inputs without the need for retraining. On the other hand, GPT-4o delivered the best predictive performance in the zero-shot setting among language models and showed further improvement with the integration of short-term memory. For open-source/open-weight models, Qwen 2.5 vision 32b showed the best result without using any complementary techniques. This demonstrates that strong baseline results are possible with certain architectures. While our framework is designed to be modular and applicable to any open-source model, we further evaluated additional models to examine the impact of each component. Llama 3.2 vision demonstrated substantial gains when both knowledge transfer and short-term memory were applied, achieving an MAE of 3.62 and 3.87 in both training and validation sets respectively, which is comparable to the performance of proprietary models. Gemma 3.0 exhibited a similar improvement under the same enhancements, highlighting the effectiveness of our proposed techniques across different models. In addition to predictive performance, the generated explanations reflected the intended abductive reasoning structure. As shown in Figure 2, the explanations combine a data-driven rationale highlighting historical trends and clinical indicators with hypothesis-driven reasoning based on plausible assumptions about the patient's behavior and comorbidities.

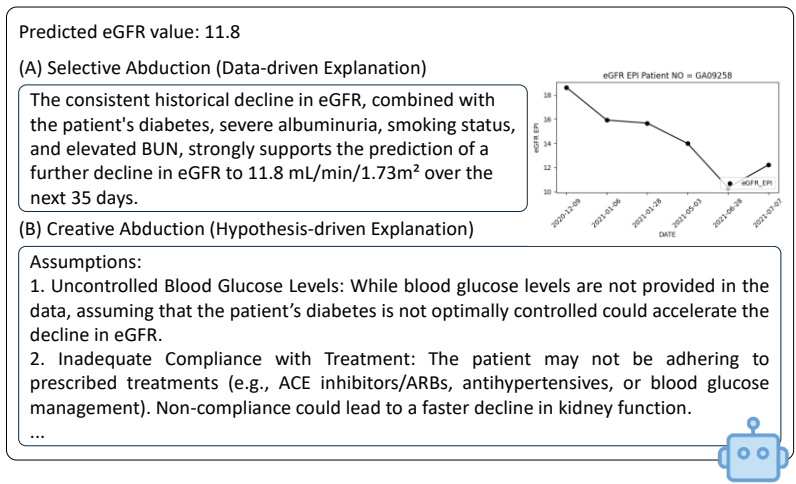

Fig. 2. An example of the prediction and explanation generated by Qwen 2.5 vision.

In this study, we demonstrated that our modular framework allows diverse LMMs to benefit from improved eGFR prediction accuracy. One of the main strengths is its flexibility, enabling targeted enhancement of weaker models without the need for retraining. This design makes the framework more accessible and easier to deploy in resource-constrained environments.

Beyond prediction accuracy, the generated explanations offer potential value for real-world applications. One immediate use case is clinical decision support, where the model's structured reasoning can help physicians better understand underlying risk factors and the rationale behind predicted outcomes. Another promising direction is medical education. This system could be used as a training tool for medical students or junior clinicians by presenting real-world clinical cases alongside AI-generated explanations. Such use may foster the capabilities of diagnostic reasoning, decision-making skills, and the ability to synthesize information from multiple clinical variables, which are critical in clinical practice.

Despite the overall effectiveness of the framework, we observed some unexpected results that warrant further consideration. For instance, Qwen 2.5 Vision 32B performed competitively without using complementary techniques. This may be attributed to stronger native visual-language integration in its architecture, allowing it to interpret eGFR trajectories without additional support. While this observation is promising, model-specific variability remains insufficiently understood and may influence the choice of enhancement strategies across different settings..

The framework also has certain limitations. It was developed and evaluated using a single dataset within a specific clinical context, which may constrain its generalizability. Furthermore, although the generated explanations are structured and clinically plausible, their practical value in real-world decision-making has yet to be validated in collaboration with domain experts. Nonetheless, the framework's modular design and integration of reasoning-based explanations offer clear potential for broader application in clinical decision support and medical training environments.

## V. CONCLUSION

We introduce a collaborative framework that enhances the predictive performance of open-source LMMs for eGFR forecasting while providing plausible explanations. The proposed framework leverages visual knowledge transfer and a short-term memory mechanism to address limitations in existing models. It also employs abductive reasoning to generate clinical explanations from observed data and plausible hypotheses. Our experimental results show that the framework achieves prediction accuracy comparable to proprietary models, while maintaining the flexibility and privacy benefits of open-source deployment. This approach provides a clinically applicable and interpretable solution for eGFR prediction, enabling accurate risk monitoring while supporting local, privacy-preserving deployment.

ACKNOWLEDGMENT

This study is mainly supported by the National Health Research Institutes (NHRI-EX114-11208PI).